\documentclass[sigconf]{acmart}

\AtBeginDocument{%
  }
\setcopyright{rightsretained}
\copyrightyear{2026}
\acmYear{2026}
\acmConference[ACM SIGSPATIAL '26]{ACM SIGSPATIAL International Conference on Advances in Geographic Information Systems}{November 2026}{Riverside, CA, USA}
\acmBooktitle{Proceedings of the 34th ACM SIGSPATIAL International Conference on Advances in Geographic Information Systems}

\begin{document}

\title[PT-WNO: Point Transformer with Wavelet Neural Operator]{PT-WNO: Point Transformer with Wavelet Neural Operator for 3D Point Cloud Semantic Segmentation}

\author{Nhut Le}
\email{nhl224@lehigh.edu}
\orcid{0009-0002-5223-0505}
\affiliation{%
  \institution{Lehigh University}
  \city{Bethlehem}
  \state{Pennsylvania}
  \country{USA}
}

\author{Maryam Rahnemoonfar}
\email{maryam@lehigh.edu}
\orcid{0000-0001-8495-9013}
\authornote{Corresponding author}
\affiliation{%
  \institution{Lehigh University}
  \city{Bethlehem}
  \state{Pennsylvania}
  \country{USA}
}

\renewcommand{\shortauthors}{Le, et al.}

\begin{abstract}
Point cloud semantic segmentation requires architectures that capture both fine-grained local geometry and broad global scene structure. Transformer-based networks have demonstrated strong performance by focusing on detailed local feature aggregation; however, global context is conveyed primarily through skip connections across encoder-decoder stages, which we argue is insufficient for full scene understanding. We hypothesize that augmenting skip connections with a learnable global feature extraction module allows the network to acquire scene-level knowledge before descending into local detail, leading to richer and more contextually grounded representations. To this end, we propose  Point Transformer with Wavelet Neural Operato (PT-WNO), which integrates a shared Wavelet Neural Operator (WNO) branch alongside the skip connections of a point cloud transformer backbone. At each encoder-decoder transition, point features are projected onto a dense 3D volumetric grid where the WNO captures multi-scale global spectral context through learnable wavelet decomposition and reconstruction. These global features are fused back into the network via lightweight adapters, complementing rather than replacing the existing skip connections. Experiments on four large-scale 3D point cloud benchmarks demonstrate the effectiveness of PT-WNO. On S3DIS (Area~5), PT-WNO achieves $71.59\%$ mIoU, outperforming the Point Transformer v3 (PTv3) baseline by $+1.03$ points. On DALES it achieves $81.05\%$ mIoU ($+1.47$ over the baseline). On ScanNet~v2, PT-WNO obtains $76.19\%$ mIoU, remaining competitive with the baseline ($76.36\%$).
\end{abstract}



\maketitle

\section{Introduction}
\label{sec:intro}

3D semantic segmentation of point clouds underpins many spatial applications, from indoor mapping and asset inventory to city‑scale digital twins. To process irregular 3D point clouds effectively, recent state-of-the-art approaches have adopted diverse structuring strategies. Classical and hierarchical methods~\cite{qi2017pointnet,qi2017pointnetplusplus,qian2022pointnext,zhao2021point,wu2022point} rely on explicit neighbor grouping (e.g., $k$-NN or ball query) and Farthest Point Sampling (FPS), which can be computationally expensive and difficult to scale. Voxel-based sparse convolutional networks~\cite{choy20194d,park2022fast,robert2023spt} address this with efficient hashing but often lack the flexible, content-dependent receptive field of transformers. Consequently, the field has recently shifted toward serialized point transformers~\cite{wu2024ptv3}, which map 3D points into 1D ordered sequences using space-filling curves (e.g., Morton or Hilbert codes). This serialization enables highly efficient, contiguous windowed attention without the latency of complex grouping or hash map construction.

While these models have strong local reasoning capabilities, their notion of global context remains implicit. In transformer-based architectures~\cite{zhao2021point,wu2022point,wu2024ptv3,park2022fast,hu2019randla}, information is propagated through stacked local attention, hierarchical pooling, and encoder–decoder skip connections. This design is effective at sharing context within and across scales, but it does not explicitly model the scene as a global operator acting over the entire spatial domain. As a result, capturing long-range structure - such as relationships between distant rooms, corridors, or large objects-still relies on many layers of local interactions, which can be suboptimal for large-scale indoor and LiDAR scenes.

In parallel, the neural operator literature has shown that treating data as functions and learning operators between function spaces can be highly effective for modeling global and multiscale phenomena~\cite{kovachki2023neural,li2021fourier}. Fourier Neural Operators (FNOs) parameterize integral kernels in the Fourier domain to enable global interactions at $\mathcal{O}(N \log N)$ cost, and have been successfully applied to parametric partial differential equations~\cite{li2021fourier}. Wavelet Neural Operators (WNOs) extend this idea by working in the wavelet domain, leveraging joint spatial–frequency localization to better capture non-stationary and scale-varying structure~\cite{tripura2022wavelet}. Despite their promise, these operator-learning techniques have been largely confined to scientific computing and have not been systematically integrated with modern point-based transformers for 3D semantic segmentation.

We argue that 3D scene understanding can benefit from combining the strengths of both paradigms: the geometric flexibility and locality of point transformers, and the global, multiscale perspective of neural operators. To this end, we present the \textit{Point Transformer with Wavelet Neural Operator (PT-WNO)}, a framework that augments a transformer-style backbone with a shared wavelet neural operator branch. PT-WNO projects intermediate point features onto a regular 3D grid, applies a wavelet neural operator to aggregate global and multiscale context, and fuses the resulting representations back into the transformer hierarchy at multiple stages. This design treats scene-level context as an explicit operator, rather than an emergent property of stacked local attention, while preserving the efficiency of serialization-based processing.

To validate our approach, we conduct experiments on four standard 3D semantic segmentation benchmarks: S3DIS~\cite{armeni2016s3dis}, ScanNet~v2~\cite{dai2017scannet}, and DALES~\cite{varney2020dales}. PT-WNO consistently improves over the strong PTv3~\cite{wu2024ptv3} baseline on three of four benchmarks, achieving gains of $+1.03$ mIoU on S3DIS (Area~5), and $+1.47$ mIoU on DALES, while remaining competitive on ScanNet~v2. These results demonstrate that incorporating wavelet neural operators into the point transformer backbone provides a consistent and generalizable benefit across diverse scene types - indoor, outdoor, aerial, and autonomous driving.

\subsection{Contributions}
This work makes the following contributions:
\begin{enumerate}
    \item \textbf{PT-WNO Architecture.} We propose the Point Transformer with Wavelet Neural Operator (PT-WNO), which augments a serialization-based point transformer backbone with a shared wavelet operator branch. This provides an explicit, efficient mechanism for modeling global and multiscale scene context over a regular 3D grid.
    \item \textbf{Operator-Based Global Context with Geometry Encoding.} We introduce a novel integration scheme that treats global context as an operator mapping over the entire scene. Unlike standard models relying solely on stacked local attention and pooling, our approach incorporates relative position encoding prior to gridding, preserving explicit geometric relationships within the global operator space.
    \item \textbf{Generalizability to Outdoor LiDAR Environments.} We validate PT-WNO on large-scale, appearance-poor outdoor sensing settings where geometric understanding is critical. Beyond indoor scenes, PT-WNO achieves notable performance gains on DALES ($+1.47$ mIoU overall, demonstrating that learning global geometric mappings effectively compensates for the lack of RGB features.
    \item \textbf{Ablation on Operator Capacity.} We conduct targeted ablation studies on the input modalities. These studies demonstrate that a single, shared universal operator branch achieves a highly favorable accuracy-efficiency trade-off for large-scale spatial applications.
\end{enumerate}
\section{Related Work}
\label{sec:related}
\begin{figure*}[ht]
  \centering
  \includegraphics[width=\linewidth]{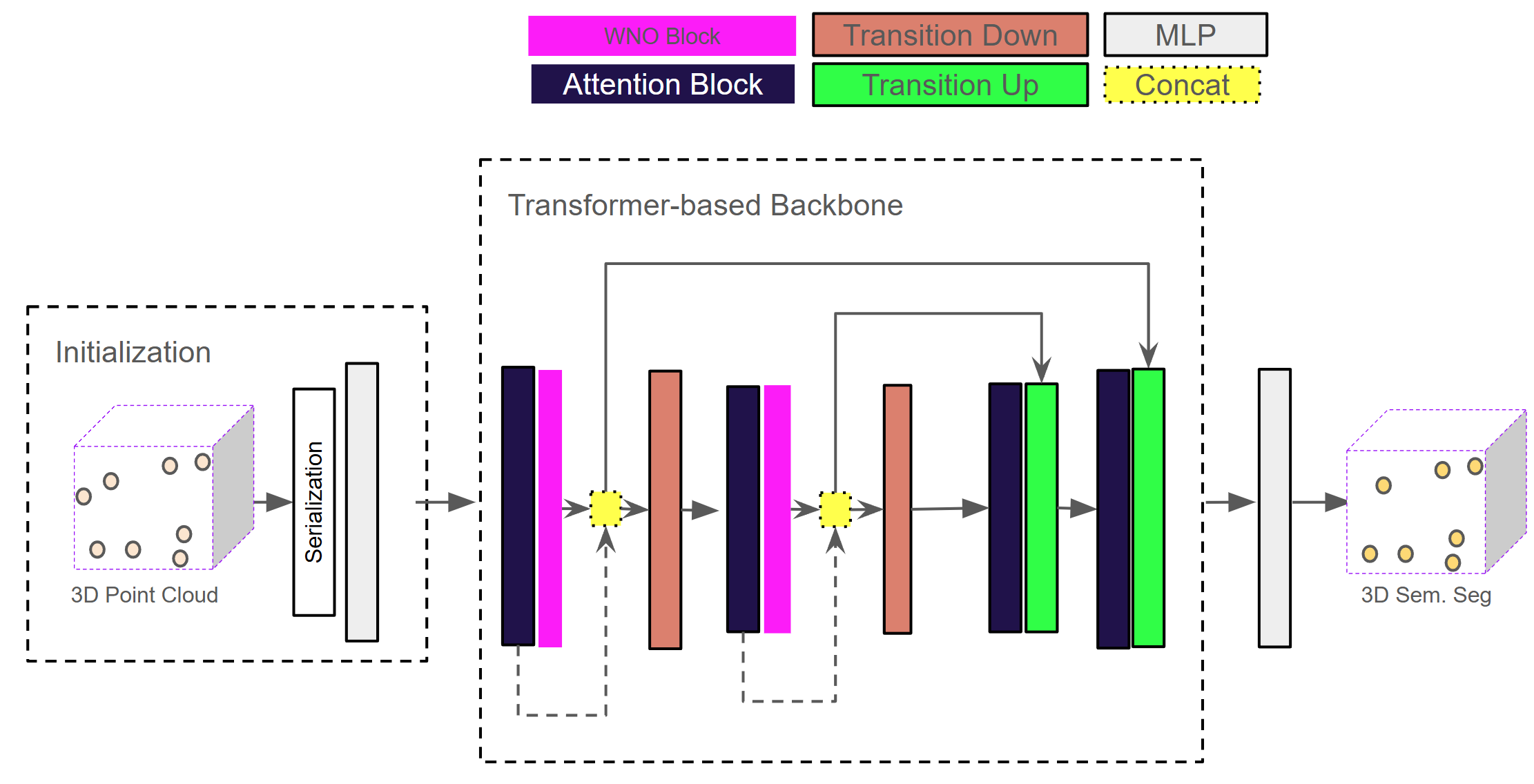}
  \caption{Architecture Overview of PT-WNO - The network integrates a localized Transformer-based backbone with a continuous Wavelet Neural Operator (WNO) block to jointly capture fine-grained geometry and global spatial layout.}
  \Description{}
    \label{fig:framework}
\end{figure*}
We review the literature relevant to PT-WNO across three primary areas: 3D semantic segmentation frameworks, neural operator learning, and explicit global context modeling in point clouds.

\subsection{3D Semantic Segmentation}
Deep learning approaches for 3D semantic segmentation broadly fall into four categories: point-based, voxel-based, graph-based, and serialization-based methods.

\subsubsection{Point, Voxel, and Graph Methods}
Point-based methods process raw point clouds directly to preserve fine-grained geometric detail. Early multi-layer perceptron (MLP)-based hierarchies~\cite{qi2017pointnet,qi2017pointnetplusplus,qian2022pointnext} laid the groundwork for local feature learning but relied on computationally expensive farthest point sampling (FPS) and $k$-NN grouping. To mitigate these bottlenecks, continuous convolutional variants like KPConv~\cite{Thomas_2019_ICCV} adapted standard kernels to irregular spatial distributions, while architectures like RandLA-Net~\cite{hu2019randla} leveraged random sampling for extreme efficiency at the cost of geometric fidelity. More recently, adaptive receptive field methods such as OA-CNNs~\cite{Peng_2024_CVPR} have advanced point-based representation capacity but still face notable scalability challenges in massive environments. Conversely, voxel-based methods~\cite{choy20194d,park2022fast} discretize continuous points into regular sparse grids to exploit highly efficient sparse convolutions, though they inevitably introduce quantization artifacts. Graph-based approaches~\cite{Landrieu2017-hv} build geometrically homogeneous superpoint graphs to model long-range structural dependencies, though they often suffer from complex, dataset-dependent graph construction overhead.

\subsubsection{Point Transformers and Serialization}
Attention-based architectures have emerged as a dominant paradigm due to their dynamic, content-dependent receptive fields. Early point transformers~\cite{zhao2021point,wu2022point} utilized vector self-attention mechanisms but were heavily constrained by localized neighbor grouping. Point Transformer V3 (PTv3)~\cite{wu2024ptv3} overcame these scalability barriers by serializing irregular 3D point sets into 1D structured sequences via static space-filling curves (e.g., Hilbert or Z-order codes), enabling highly efficient, contiguous windowed attention that operates at linear complexity. Consequently, PTv3~\cite{wu2024ptv3} achieves state-of-the-art performance across major indoor benchmarks like the Stanford Large-Scale 3D Indoor Spaces (S3DIS)~\cite{armeni2016s3dis} and ScanNet v2~\cite{dai2017scannet}, as well as outdoor datasets~\cite{Caesar_2020_CVPR,behley2019semantickitti,Sun_2020_CVPR}. However, because windowed attention limits the receptive field to localized serialized segments, global scene context remains largely implicit and unmodeled at a holistic level. Furthermore, space-filling curves can inadvertently separate spatially adjacent points in the 1D traversal sequence, a limitation that motivates an explicit, spatial-domain global branch.
\subsection{Neural Operators}
Neural operators represent a paradigm shift by learning mappings between infinite-dimensional function spaces rather than finite-dimensional vectors, rendering them fundamentally resolution-independent and highly effective at capturing macro-scale system behaviors~\cite{kovachki2023neural,li2021fourier}. The Fourier Neural Operator (FNO)~\cite{li2021fourier} parameterizes integral kernels in the frequency domain via the Fast Fourier Transform (FFT), enabling global feature interactions at an elegant $\mathcal{O}(N \log N)$ computational cost. While highly successful in solving partial differential equations (PDEs), FNO struggles with localized, non-periodic sharp boundaries due to the global extension of the Fourier basis. 

The Wavelet Neural Operator (WNO)~\cite{tripura2022wavelet} elegantly resolves this by computing learnable operations in the wavelet domain, exploiting the joint spatial-frequency localization of wavelets to seamlessly handle multi-scale and non-stationary phenomena. This dual localization is exceptionally well-suited for real-world 3D scenes captured by LiDAR or depth sensors, where sharp geometric boundaries (e.g., walls, columns, thin frames) coexist with broad spatial layouts. Historically, neural operators have been confined to scientific computing and continuous physical simulations. To the best of our knowledge, PT-WNO represents the first framework to systematically integrate a wavelet neural operator as a shared global-context branch within a serialization-based point transformer backbone for 3D semantic segmentation.

\subsection{Global Context in Point Cloud Segmentation}
Augmenting local point cloud networks with broader contextual understanding is a long-standing challenge. Indeed, capturing long-range dependencies and non-local features has been shown to significantly improve the performance of deep neural networks on point cloud segmentation by resolving local geometric ambiguities~\cite{han2020point2node, wei2020multi, deng2021ga}. To achieve this, prior works have explored dynamic node correlations~\cite{han2020point2node}, heuristic region mining~\cite{wei2020multi}, and dedicated global attention modules~\cite{deng2021ga}. While these non-local blocks and global attention mechanisms~\cite{wang2018non} successfully enable long-range interactions across an entire feature map, their quadratic complexity with respect to point sequence length makes them computationally intractable for large-scale 3D scenes. To circumvent this, contemporary point transformers rely on a combination of windowed attention, hierarchical downsampling, and encoder-decoder skip connections~\cite{zhao2021point,wu2022point,wu2024ptv3} to progressively propagate contextual information across scales. While effective, this setup means that "global" context is merely an emergent property of deeply stacked local operations rather than an explicit, scene-level representation. 

More recently, alternative sequence-modeling frameworks, such as State Space Models (SSMs) and Mamba variants~\cite{gu2024mamba}, have been explored to capture long-range dependencies with linear complexity by scanning across serialized point sequences. However, these sequential methods remain highly sensitive to point ordering heuristics. PT-WNO introduces a fundamentally different perspective: instead of relying on sequential scanning or deep attention hierarchies alone, we project intermediate multi-scale point features onto a shared, regular 3D grid and apply a continuous wavelet operator. This branch performs explicit global and multi-scale spectral aggregation, injecting holistic scene structure directly into multiple stages of the transformer backbone. This dual-pathway approach proves highly advantageous for both complex indoor layouts and color-sparse outdoor LiDAR analytics.
\begin{table}[ht]
\centering
\caption{Semantic segmentation results (mIoU \%) across benchmarks. Gray values are from original publications; reproduced baselines and PT-WNO variants are run under identical constraints.}
\label{tab:performance}
\begin{tabular}{l c c c}
\toprule
\textbf{Method} & S3DIS (Area 5) & DALES & Scannet v2\\
\midrule
PointNet++~\cite{qi2017pointnetplusplus} & - & 68.3 & 53.5\\
SPG~\cite{Landrieu2017-hv} & 58 & 60.6 & -\\
ConvPoint~\cite{boulch2020convpoint} & 68.2 & 67.4 & -\\
KPConv~\cite{Thomas_2019_ICCV} & 67.1 & \textbf{81.1} & 69.2\\
MinkUNet~\cite{choy20194d} & 65.4 & - &72.2\\
SPT~\cite{robert2023spt} &68.9 & 79.6 & -\\
\midrule
PTv3~\cite{wu2024ptv3} & {\color{gray}74.70} & {\color{gray}-} & {\color{gray} 77.5}\\
\rotatebox[origin=c]{180}{$\Lsh$} \textit{reproduced} & 70.56 & 79.58 & \textbf{76.36}\\
\textbf{PT-WNO (ours)} & \textbf{71.59} & 81.05 & 76.19\\
\bottomrule
\end{tabular}%
\end{table}

\begin{table*}[t]
\centering
\caption{Class-wise IoU (\%) on S3DIS Area 5.}
\label{tab:s3dis_classwise}
\resizebox{\linewidth}{!}{%
\begin{tabular}{l c|c|c| c c c c c c c c c c c c c}
\toprule
\textbf{Method} &\textbf{OA} & \textbf{mAcc} &\textbf{mIoU}  & \rotatebox{90}{ceiling} & \rotatebox{90}{floor} & \rotatebox{90}{wall} & \rotatebox{90}{beam} & \rotatebox{90}{column} & \rotatebox{90}{window} & \rotatebox{90}{door} & \rotatebox{90}{table} & \rotatebox{90}{chair} & \rotatebox{90}{sofa} & \rotatebox{90}{bookcase} & \rotatebox{90}{board} & \rotatebox{90}{clutter} \\
\midrule
Ptv1~\cite{zhao2021point} & 90.8 &	76.5&	70.4&	\textbf{94}&	\textbf{98.5}	&\textbf{86.3}	&0	& 38&	\textbf{63.4}&	74.3&	\textbf{89.1}&	82.4&	74.3&	\textbf{80.2}&	76&	59.3\\
\midrule
PTv3~\cite{wu2024ptv3} (reproduced) & 91.04 & 75.56	&70.56& 93.73 & 98.35 & 84.92 & 0.00 & 35.50 & 61.87 & 72.69 & 82.19 & 91.45 & 68.51 & 79.18 & \textbf{85.18} & \textbf{63.70} \\
\midrule
PT-WNO (ours) &\textbf{91.28}&	\textbf{76.9}	&\textbf{71.59} & 93.91 & 98.35 & 85.75  & 0.00 & 40.27 & 61.89 & \textbf{75.09} & 83.74 & 92.63 & 75.72 & 78.1 & 82.67 & 62.52 \\
\rotatebox[origin=c]{180}{$\Lsh$} \textit{w/o color feature} &	90.87&	77.21&	70.04&	93.90&	98.37&	85.29&	0.00&	\textbf{46.94}&	62.04&	71.72&	81.50&	\textbf{93.16}&	\textbf{78.10}&	76.64&	61.14&	61.77\\
\bottomrule
\end{tabular}%
}
\end{table*}

\begin{table*}[t]
\centering
\caption{Class-wise IoU (\%) on ScanNet v2 (validation). Only the controlled PTv3 baseline and PT-WNO are compared to isolate the effect of the wavelet operator.}
\label{tab:scannet_classwise}
\resizebox{\linewidth}{!}{%
\begin{tabular}{l | c | c c c c c c c c c c c c c c c c c c c c}
\toprule
\textbf{Method} & \textbf{mIoU} & \rotatebox{90}{wall} & \rotatebox{90}{floor} & \rotatebox{90}{cabinet} & \rotatebox{90}{bed} & \rotatebox{90}{chair} & \rotatebox{90}{sofa} & \rotatebox{90}{table} & \rotatebox{90}{door} & \rotatebox{90}{window} & \rotatebox{90}{bookshelf} & \rotatebox{90}{picture} & \rotatebox{90}{counter} & \rotatebox{90}{desk} & \rotatebox{90}{curtain} & \rotatebox{90}{refriger.} & \rotatebox{90}{shower} & \rotatebox{90}{toilet} & \rotatebox{90}{sink} & \rotatebox{90}{bathtub} & \rotatebox{90}{otherfurn.} \\
\midrule
PTv3~\cite{wu2024ptv3} (reproduced) & \textbf{76.36} & 86.5 & 95.35 & \textbf{71.09} &  \textbf{86.03} &\textbf{ 92.37} & 83.53 &\textbf{ 77.1} & 70.38 & 68.17 & 80.46 & 37.01 & \textbf{73.15} &\textbf{ 70.1} & 77.89 & 68.1 & 71.61 & 95.81 & \textbf{70.2} & \textbf{88.88} & \textbf{63.45}\\
\textbf{PT-WNO} & 76.19 & \textbf{86.7} & \textbf{95.65} & 69.45 & 82.61 & 91.61 & \textbf{83.91} &  75.28 & \textbf{72.58} & \textbf{71.7} & \textbf{80.92} & \textbf{38.69} & 67.32 & 69.16 & \textbf{78.36} & \textbf{74.06} & \textbf{72.89} & \textbf{95.82} & 68.14 & 83.11 & 60.90\\
\bottomrule
\end{tabular}%
}
\end{table*}

\begin{table*}[h]
\centering
\caption{Semantic segmentation results (\%) (mIoU) and class-wise IoU on the DALES aerial dataset.}
\label{tab:dales_classwise}
\resizebox{\linewidth}{!}{%
\begin{tabular}{l | c | c | c c c c c c c c }
\toprule
\textbf{Method} & \textbf{mIoU} & \shortstack{\textbf{mIoU} \\ w/o ground} & ground & vegetation & cars &trucks & power lines & fences & poles & buildings \\
\midrule
PTv3~\cite{wu2024ptv3} (reproduced) & 79.58 & 70.35 & 96.08 & 92.23 & 82.93 & 40.69 & \textbf{95.86} & 60.99 & 71.28 & 96.59 \\
\textbf{PT-WNO (ours)} &  \textbf{81.05} & \textbf{72.41} & \textbf{96.53} & \textbf{96.53} & \textbf{85.30} & \textbf{41.97} & 95.81 & \textbf{64.11} &\textbf{ 74.84} & \textbf{97.15}\\
\bottomrule
\end{tabular}%
}
\end{table*}
\section{PT-WNO: Point Transformer with Wavelet Neural Operator}
\label{sec:framework}
\begin{figure*}[ht]
  \centering
  \includegraphics[width=\linewidth]{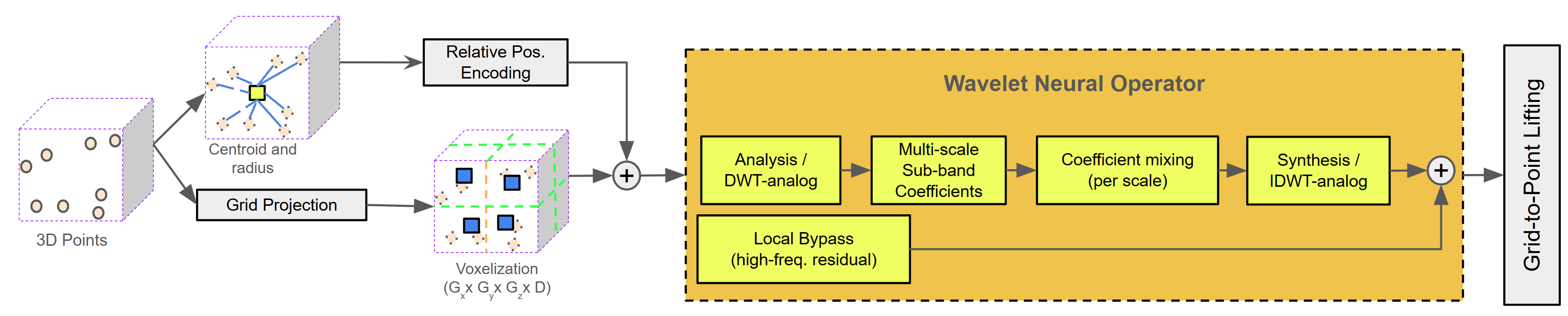}
  \caption{Detailed architecture of the WNO Block - The block performs a learned multiresolution analysis-synthesis process on a voxelized 3D grid, progressively encoding features to coarser resolutions, applying scale-wise feature mixing and bottleneck processing, and reconstructing the grid through residual fusion to produce globally contextualized features.}
  \label{fig:wno}
\end{figure*}
Let a 3D scene be represented as a continuous point set $\mathcal{P} = \{p_i\}_{i=1}^{N}$, where each point is associated with a raw feature vector $\mathbf{x}_i = [\mathbf{c}_i \,\Vert\, \mathbf{a}_i] \in \mathbb{R}^{F}$ comprising its 3D spatial coordinates $\mathbf{c}_i \in \mathbb{R}^3$ and auxiliary attributes $\mathbf{a}_i$ (such as color or LiDAR reflection intensity). The fundamental objective of 3D semantic segmentation is to map this input to a dense prediction space, assigning a discrete semantic label $y_i \in \{1, \dots, C\}$ to every point in $\mathcal{P}$.

To overcome the inherent trade-off between fine-grained local geometric modeling and large-scale scene understanding, we introduce \textbf{PT-WNO} (Point Transformer with Wavelet Neural Operator). As illustrated in Figure~\ref{fig:framework}, PT-WNO adopts a powerful architecture that processes the point cloud concurrently in two distinct topological spaces:
\begin{itemize}
    \item \textbf{Transformer-Based Backbone:} Operates directly on the irregular 3D point cloud. Serialized space-filling curves and windowed self-attention efficiently capture high-resolution, localized geometric features without the memory overhead of dense 3D grids.

    \item \textbf{Wavelet Neural Operator Block:} Operates on a regularized, voxelized grid. It explicitly models continuous global context by learning an integral operator in the multi-scale wavelet domain, capturing long-range structural dependencies such as room-scale wall layouts or outdoor road corridors.
\end{itemize}

\subsection{Pipeline Overview} The forward pass of PT-WNO proceeds through a unified, end-to-end pipeline. First, the raw point cloud $\mathcal{P}$ is serialized into a structured 1D sequence via a space-filling curve (e.g., Hilbert or Z-order), imposing a consistent traversal order that enables efficient windowed attention in subsequent stages. An input MLP then projects the raw features $\mathbf{x}_i$ of each serialized point into a high-dimensional initial embedding. These embeddings are passed through a U-Net-style encoder hierarchy. At multiple stages within the encoder, the localized point features are temporarily projected (voxelized) onto a dense 3D grid and enriched with relative positional encodings. The shared WNO block processes this grid, performing a learned multiresolution analysis to mix features across spatial scales. The resulting globally contextualized features are immediately lifted back to the sparse point domain and fused with the local point features via residual concatenation. Finally, a point-wise decoder restores the original spatial resolution, outputting the final
semantic predictions optimized by a combined Cross-Entropy and Lovász-Softmax loss.

The detailed mechanics of the backbone, the continuous wavelet operator, and the multi-stage fusion strategy are formulated in the following subsections.

\subsection{Transformer-based Backbone}

The backbone follows a U-Net-style encoder-decoder architecture. Each input point $p_i$ is associated with a raw feature vector $\mathbf{x}_i = [\mathbf{c}_i \,\Vert\, \mathbf{a}_i] \in \mathbb{R}^{F}$, where $\mathbf{c}_i \in \mathbb{R}^3$ denotes the 3D coordinates and $\mathbf{a}_i$ denotes the auxiliary attributes (e.g., color, normals). An input MLP $\psi$ projects each raw feature into a high-dimensional embedding:
\begin{equation}
    \mathbf{f}_i^{(0)} = \psi(\mathbf{x}_i) \in \mathbb{R}^{d_0},
    \label{eq:embedding}
\end{equation}
where $d_0$ is the base channel dimension. The encoder (Enc) then applies \emph{serialized windowed self-attention} blocks, where points are grouped into local windows using a space-filling curve (e.g., Hilbert or Z-order), reducing neighbor-search complexity from $O(N \log N)$ to $O(N)$~\cite{wu2024ptv3}. At encoder stage $s$, the backbone produces a downsampled point set $\mathcal{P}^{(s)}$ with per-point features:
\begin{equation}
    \mathbf{h}_i^{(s)} = \text{Enc}^{(s)}\!\left(\mathbf{f}_i^{(s-1)}\right) \in \mathbb{R}^{d_s},
    \label{eq:encoder}
\end{equation}
where $d_s$ increases with depth as spatial resolution decreases. Between encoder stages, a Grid Pooling layer (Transition Down) merges points within the same voxel to reduce spatial resolution while increasing channel depth. The decoder restores resolution via linear interpolation (Transition Up), with encoder skip connections preserving fine-grained geometric detail. While this backbone provides strong local and multiscale reasoning, global context emerges only implicitly through depth, pooling, and skip connections. The WNO branch directly addresses this limitation.

\subsection{Wavelet Neural Operator Block}
\label{sec:wno}
The WNO branch treats the backbone features $\mathbf{h}_i^{(s)}$ at each active encoder stage $s$ as samples of a continuous function over 3D space, and learns a global operator mapping them to a globally contextualized representation~\cite{tripura2022wavelet,li2021fourier}. It consists of four steps: grid projection, relative position encoding, wavelet neural operator, and grid-to-point lifting as illustrated in Figure~\ref{fig:wno}.

\subsubsection{Grid Projection.}
At each active encoder stage $s$, the backbone features $\mathbf{h}_i^{(s)} \in \mathbb{R}^{d_s}$ are first projected into the shared operator space via a per-stage lightweight adapter:
\begin{equation}
    \hat{\mathbf{h}}_i^{(s)} = \text{MLP}_{\text{in}}^{(s)}\!\left(\mathbf{h}_i^{(s)}\right)
    \in \mathbb{R}^{D},
    \label{eq:proj_in}
\end{equation}
where $D$ is the fixed universal channel dimension shared across all stages. The projected features $\hat{\mathbf{h}}_i^{(s)}$ are then mapped onto a fixed regular grid of resolution $G_x \times G_y \times G_z$. Each point coordinate $\mathbf{q}_i$ is mapped to a discrete voxel index via a normalized linear mapping:
\begin{equation}
    v_k(i) = \left\lfloor
        \frac{q_{i,k} - \min_j q_{j,k}}{\max_j q_{j,k} - \min_j q_{j,k} + \epsilon}
        \cdot (G_k - 1)
    \right\rfloor, \quad k \in \{x,y,z\},
    \label{eq:voxelize}
\end{equation}
clipped to $[0, G_k-1]$. Multiple points falling into the same voxel are aggregated via mean pooling:
\begin{equation}
    \mathbf{U}^{(s)}[v] = \frac{1}{|\{i:\,v(i)=v\}|}
    \sum_{i:\,v(i)=v} \hat{\mathbf{h}}_i^{(s)} \in \mathbb{R}^{D},
    \label{eq:grid}
\end{equation}
and unoccupied voxels are zero-filled, yielding the dense spatial tensor
$\mathbf{U}^{(s)} \in \mathbb{R}^{G_x \times G_y \times G_z \times D}$.

\subsubsection{Relative Position Encoding.}
Since the dense grid loses explicit spatial awareness after voxelization, we enrich each point feature with a relative position encoding before voxelization. Using the integer grid coordinates $\mathbf{q}_i \in \mathbb{Z}^3$ of each point, we compute the displacement from the per-stage centroid $\bar{\mathbf{q}}^{(s)} = \frac{1}{N_s}\sum_i \mathbf{q}_i^{(s)}$ and its radius $r_i = \|\boldsymbol{\delta}_i\|_2$, where $\boldsymbol{\delta}_i = \mathbf{q}_i - \bar{\mathbf{q}}^{(s)}$. Each axis is normalized by its maximum absolute displacement and the radius by its global maximum:
\begin{equation}
    \tilde{\delta}_{i,k} = \frac{\delta_{i,k}}{\max_i |\delta_{i,k}| + \epsilon},\
    k \in \{x,y,z\}, \qquad
    \tilde{r}_i = \frac{r_i}{\max_i r_i + \epsilon}.
    \label{eq:normalize}
\end{equation}
The four normalized scalars are projected via a small MLP with LayerNorm and GELU activation:
\begin{equation}
    \mathbf{e}_i = \text{MLP}_{\text{pos}}\!\left(
        [\tilde{\delta}_{i,x},\, \tilde{\delta}_{i,y},\, \tilde{\delta}_{i,z},\, \tilde{r}_i]
    \right) \in \mathbb{R}^{D},
    \label{eq:rpe}
\end{equation}
and fused point-wise into the projected feature before voxelization: $\tilde{\mathbf{h}}_i^{(s)} = \hat{\mathbf{h}}_i^{(s)} + \mathbf{e}_i$. This normalization ensures scale-invariance across scenes of different physical extents, enabling the shared operator $\mathcal{K}_\theta$ to reason about where features occur in the scene, not just what they are.

\subsubsection{Wavelet Neural Operator.}
The core operator $\mathcal{K}_\theta$ is grounded in the Wavelet Neural Operator (WNO) framework~\cite{tripura2022wavelet}, which learns a kernel integral operator by diagonalizing it in the wavelet domain. Formally, given the enriched grid $\tilde{\mathbf{U}}^{(s)}$ treated as a discretization of a continuous function $u: \mathbb{R}^3 \to \mathbb{R}^D$, the ideal WNO layer applies a 3D Discrete Wavelet Transform (DWT), mixes the resulting approximation and detail sub-band coefficients via learnable $1{\times}1{\times}1$ convolutions $R_\theta$, and reconstructs the spatial output via the Inverse DWT (IDWT), with a local bypass term $B$ to preserve high-frequency detail:
\begin{align}
\begin{split}
    \mathbf{V}^{(s)} &= \mathcal{K}_\theta\!\left(\tilde{\mathbf{U}}^{(s)}\right) \\
    &= \text{IDWT}\!\left(R_\theta \odot
       \text{DWT}\!\left(\tilde{\mathbf{U}}^{(s)}\right)\right)
       + B\!\left(\tilde{\mathbf{U}}^{(s)}\right).
    \label{eq:wno_ideal}
\end{split}
\end{align}

The original WNO~\cite{tripura2022wavelet} realizes this via fixed wavelet filter banks (e.g., Daubechies wavelets), explicitly decomposing the input into approximation and detail sub-bands before applying learnable $1{\times}1$ coefficient mixing and a fixed IDWT for reconstruction. In our setting, however, fixed filter banks introduce two practical limitations: (i)~standard wavelet libraries are designed for regular 1D/2D signals and do not extend trivially to the sparse, irregular 3D geometry of point cloud features; and (ii)~fixed filters cannot adapt to the task-specific frequency structure of
3D scene understanding~\cite{liu2018multi,mallat1989theory}. We therefore replace the fixed DWT/IDWT with fully \emph{trainable} depthwise strided convolutions (analysis) and transposed convolutions (synthesis), preserving the WNO inductive bias of hierarchical multi-scale decomposition and residual reconstruction while enabling end-to-end optimization of the decomposition bases for 3D point cloud segmentation.

\subsubsection{Grid-to-Point Lifting.}
The WNO branch is applied to thepre-pooling fine-resolution features at each encoder transition. After the operator produces the output grid $\mathbf{V}^{(s)}$ (defined in Eq.~\ref{eq:wno_ideal}), each point $p_i$ retrieves its globally contextualized feature by looking up its corresponding voxel index and projecting back to the backbone channel dimension $d_s$:
\begin{equation}
    \mathbf{g}_i^{(s)} = \text{MLP}_{\text{out}}^{(s)}\!\left(\mathbf{V}^{(s)}[v(i)]\right)
    \in \mathbb{R}^{d_s},
    \label{eq:lifting}
\end{equation}
where $v(i)$ is the voxel index of point $p_i$. During encoding, the resulting fine-resolution features are further aggregated to the coarser post-pooling resolution via max-pooling over the cluster assignments: $\mathbf{g}_i^{(s)\prime} = \max_{j:\,\text{cluster}(j)=i} \mathbf{g}_j^{(s)}$.
During decoding, the operator features are broadcast back to the finer resolution via the pooling inverse index $\pi(i)$, replacing $v(i)$ with $v(\pi(i))$ in Eq.~\ref{eq:lifting}. The operator weights $\mathcal{K}_\theta$ are fully shared across all encoder and decoder stages, keeping the parameter footprint minimal regardless of network depth.

\subsection{Multi-Stage Feature Fusion and End-to-End Training}
\label{sec:fusion_and_training}
Having lifted the globally contextualized grid features back to the point domain (Section~\ref{sec:wno}), we now describe how they are integrated with the local backbone features at each encoder stage. At any active stage $s$, the backbone feature $\mathbf{h}_i^{(s)}$ and the operator feature $\mathbf{g}_i^{(s)}\prime$ for point $p_i$ are fused via concatenation, linear projection, and a non-linear activation:
\begin{equation}
    \tilde{\mathbf{h}}_i^{(s)} = \phi\!\left(\mathbf{W}^{(s)}\,
    [\mathbf{h}_i^{(s)} \,\Vert\, \mathbf{g}_i^{(s)\prime}]\right),
\end{equation}
where $\mathbf{W}^{(s)}$ is a learnable weight matrix and $\phi$ is a composite of Layer Normalization and GELU activation. The fused representation $\tilde{\mathbf{h}}_i^{(s)}$ replaces $\mathbf{h}_i^{(s)}$ in all subsequent transformer blocks, and is carried through standard U-Net skip connections into the decoder, ensuring operator-enhanced global context propagates into all high-resolution output predictions. Because fusion is independently configurable per stage, the architecture naturally supports targeted stage-wise ablations.

The entire PT-WNO model is optimized end-to-end without branch-specific auxiliary losses, as the WNO block is continuous and fully differentiable. We employ a combined Cross-Entropy and Lovász-Softmax loss~\cite{berman2018lovasz}:
\begin{equation}
    \mathcal{L}_{\text{total}} = \mathcal{L}_{\text{CE}} + \mathcal{L}_{\text{Lovász}}.
\end{equation}
$\mathcal{L}_{\text{CE}}$ supervises per-point semantic class predictions, while $\mathcal{L}_{\text{Lovász}}$ directly optimizes the mIoU surrogate, stabilizing gradients and balancing the skewed class distributions common in both indoor and outdoor 3D datasets.
\section{Experiments}
\label{sec:experiments}
\begin{table*}[ht]
\caption{Qualitative Results on S3DIS for PT-WNO}
\label{tab:qualitative_s3dis}
    \centering
\begin{tabular}{cccc}
 Input & \includegraphics[width=0.23\linewidth]{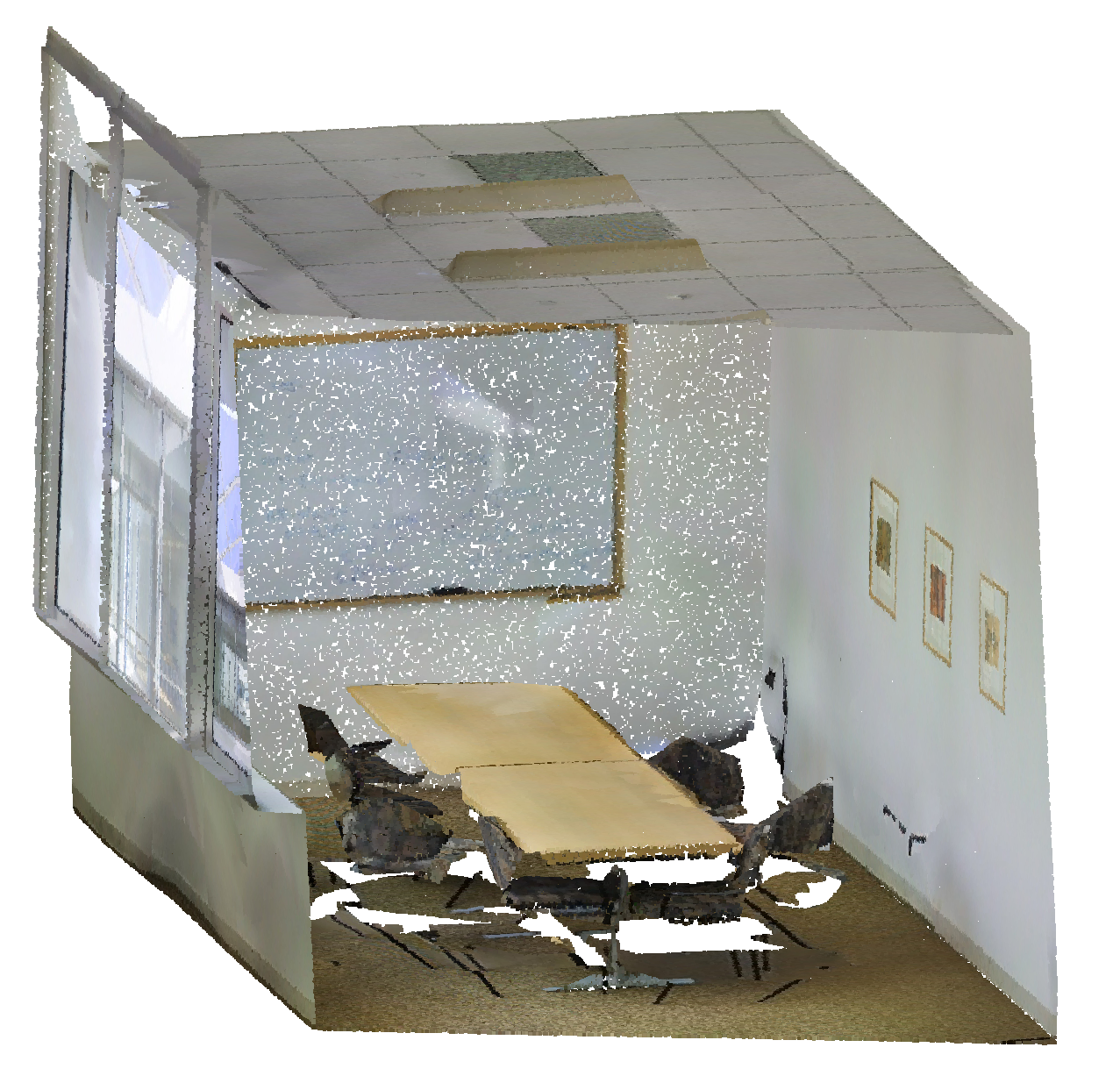}  & \includegraphics[width=0.3\linewidth]{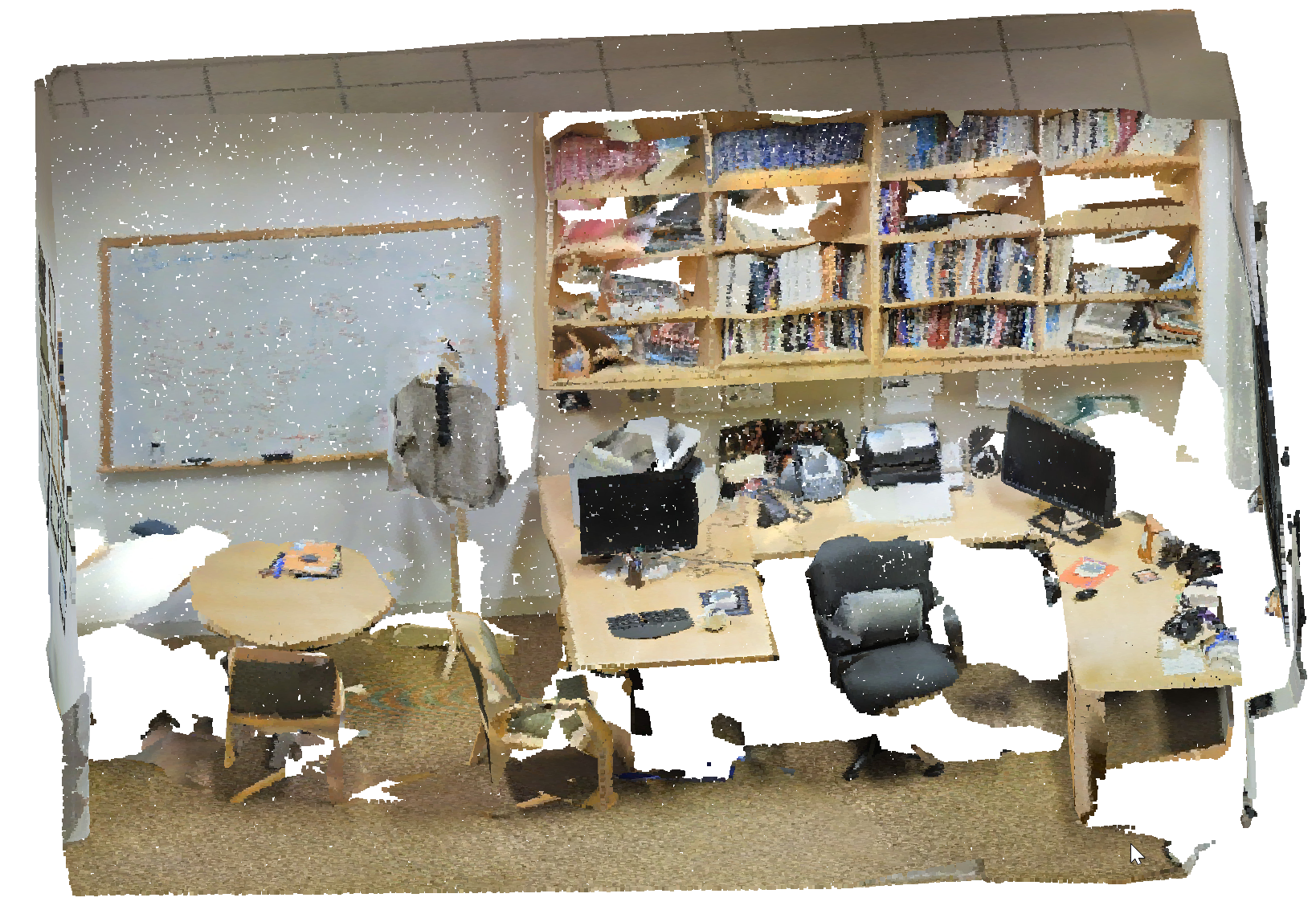} & \includegraphics[width=0.23\linewidth]{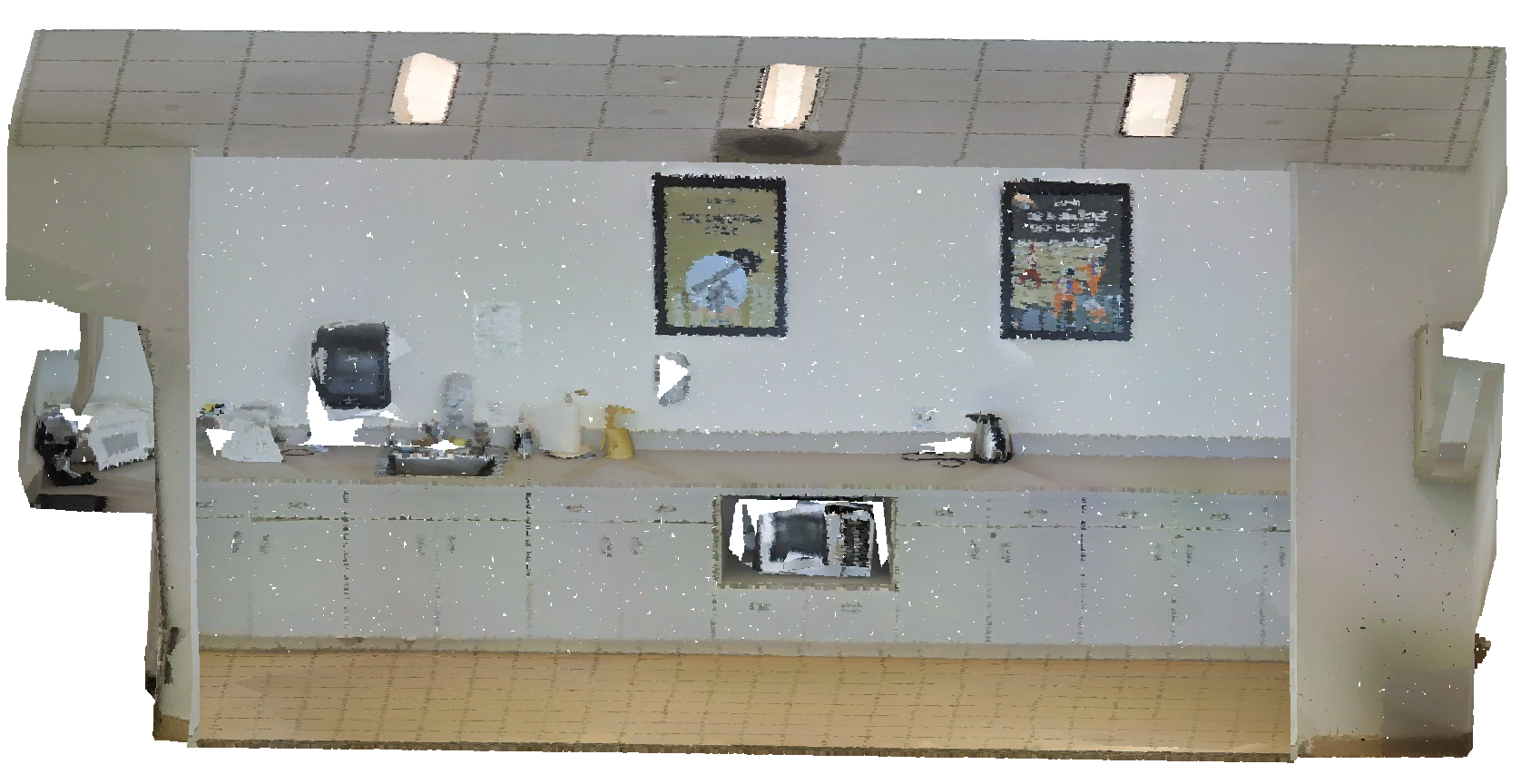} \\ 
Ground Truth & \includegraphics[width=0.23\linewidth]{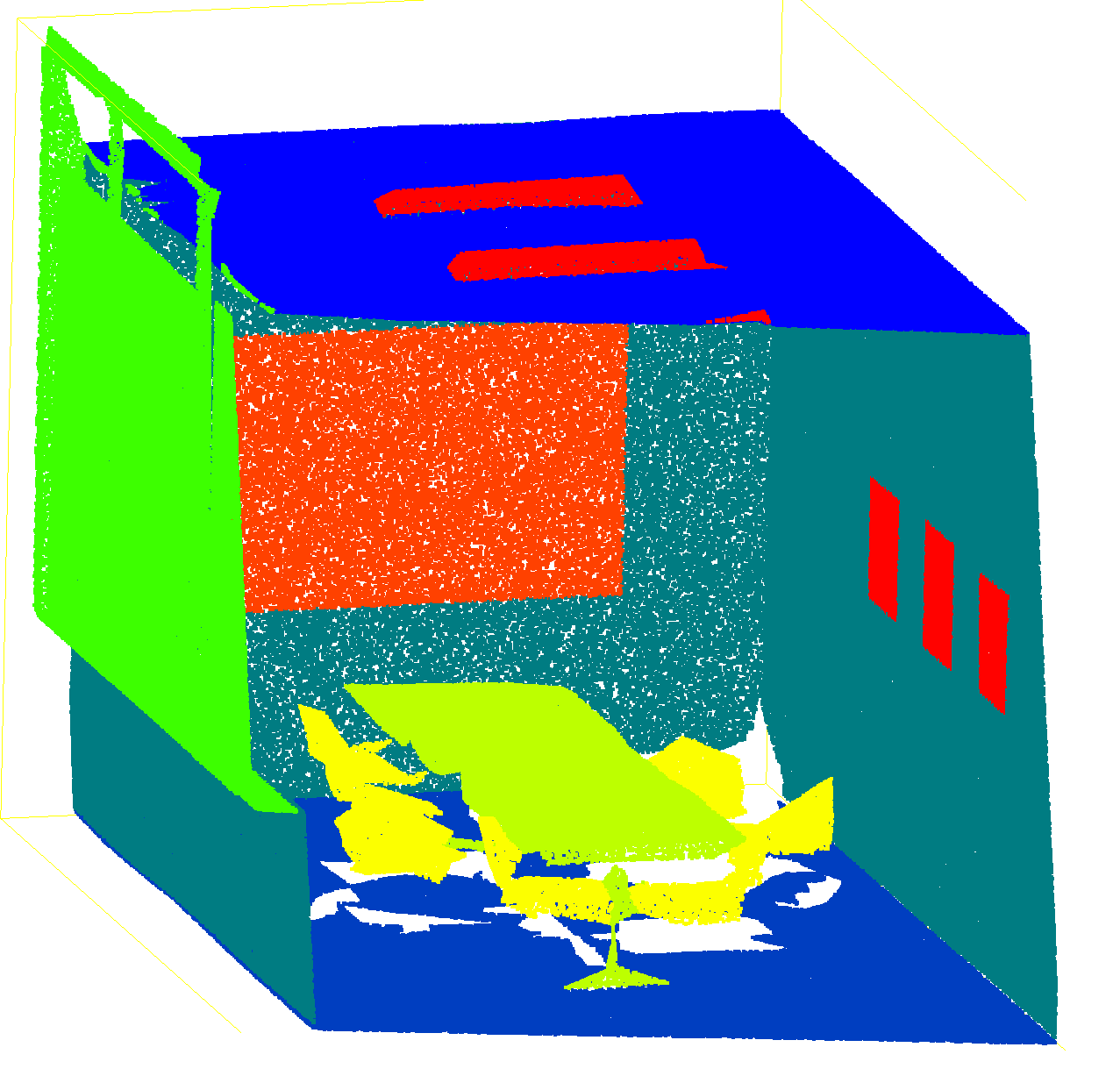}  & \includegraphics[width=0.3\linewidth]{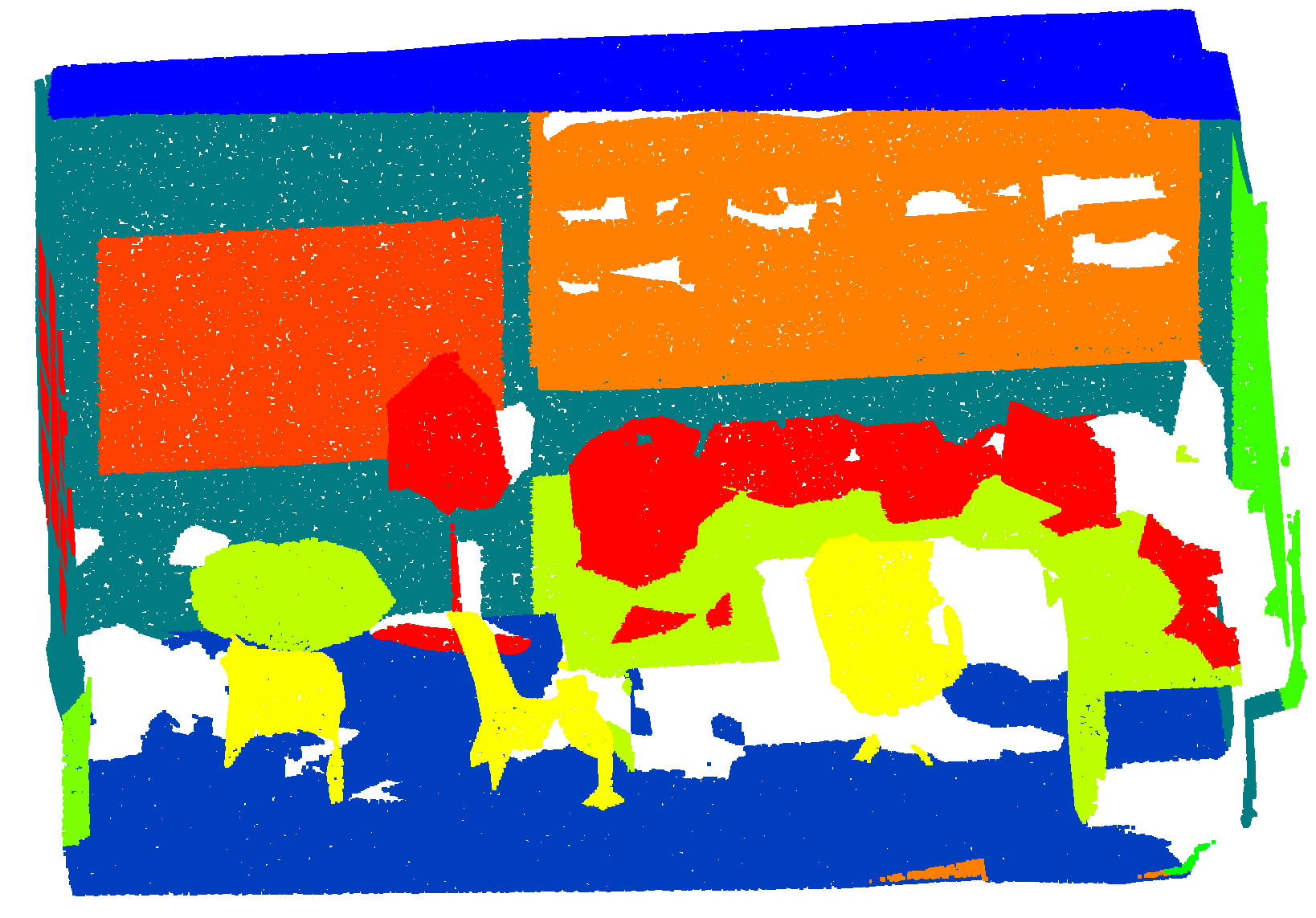} & \includegraphics[width=0.23\linewidth]{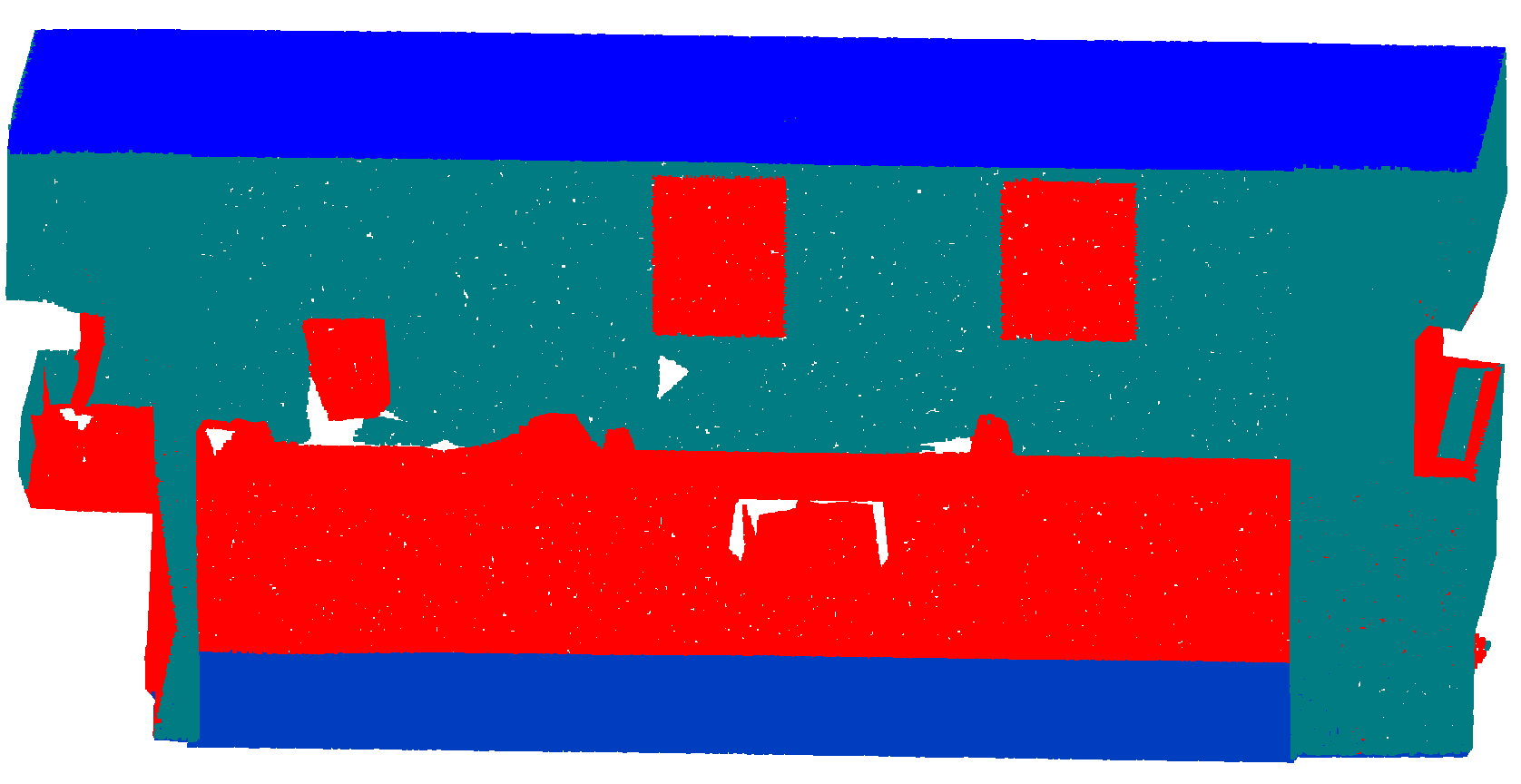}  \\
Prediction & \includegraphics[width=0.23\linewidth]{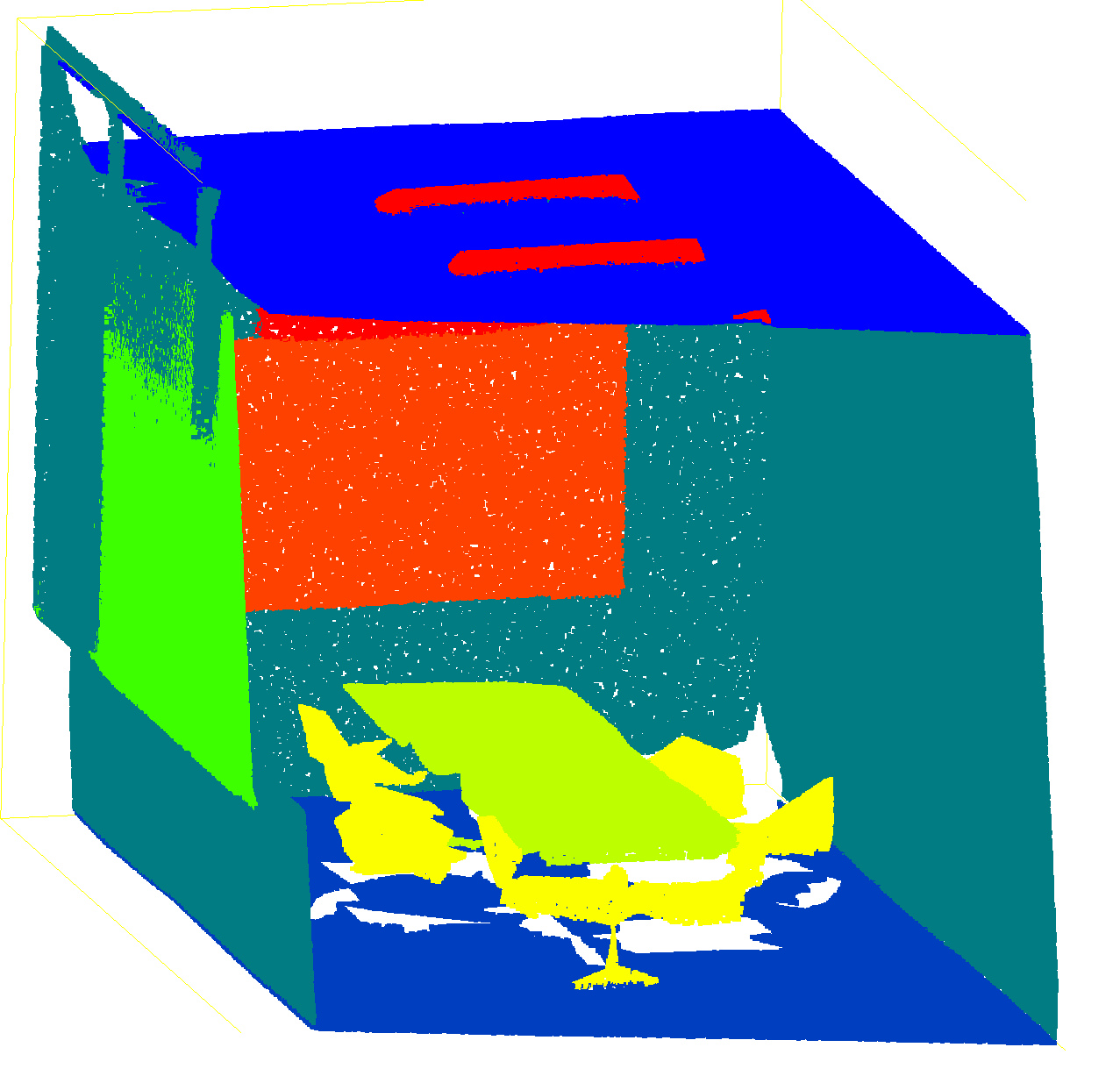}  & \includegraphics[width=0.3\linewidth]{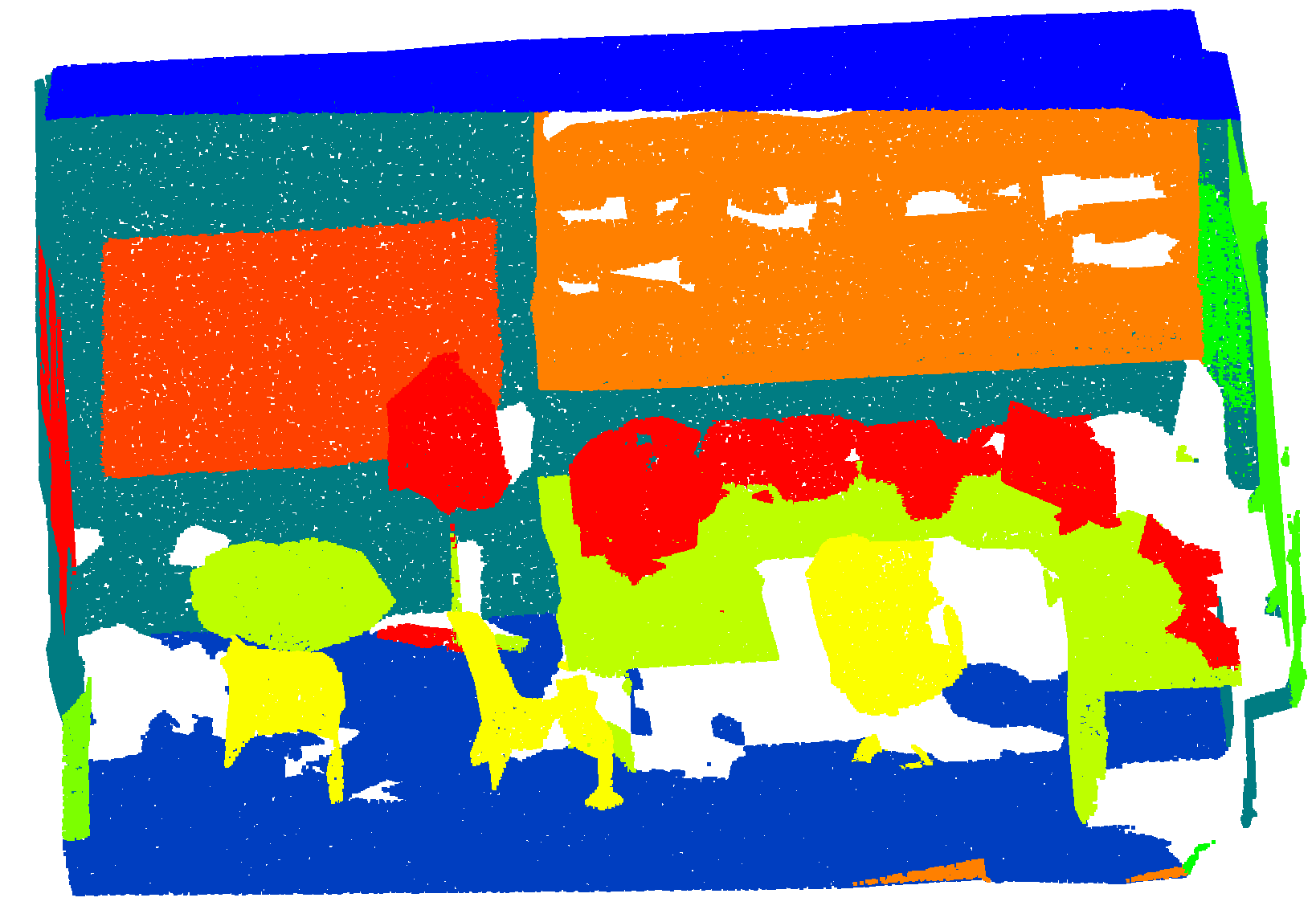} & \includegraphics[width=0.23\linewidth]{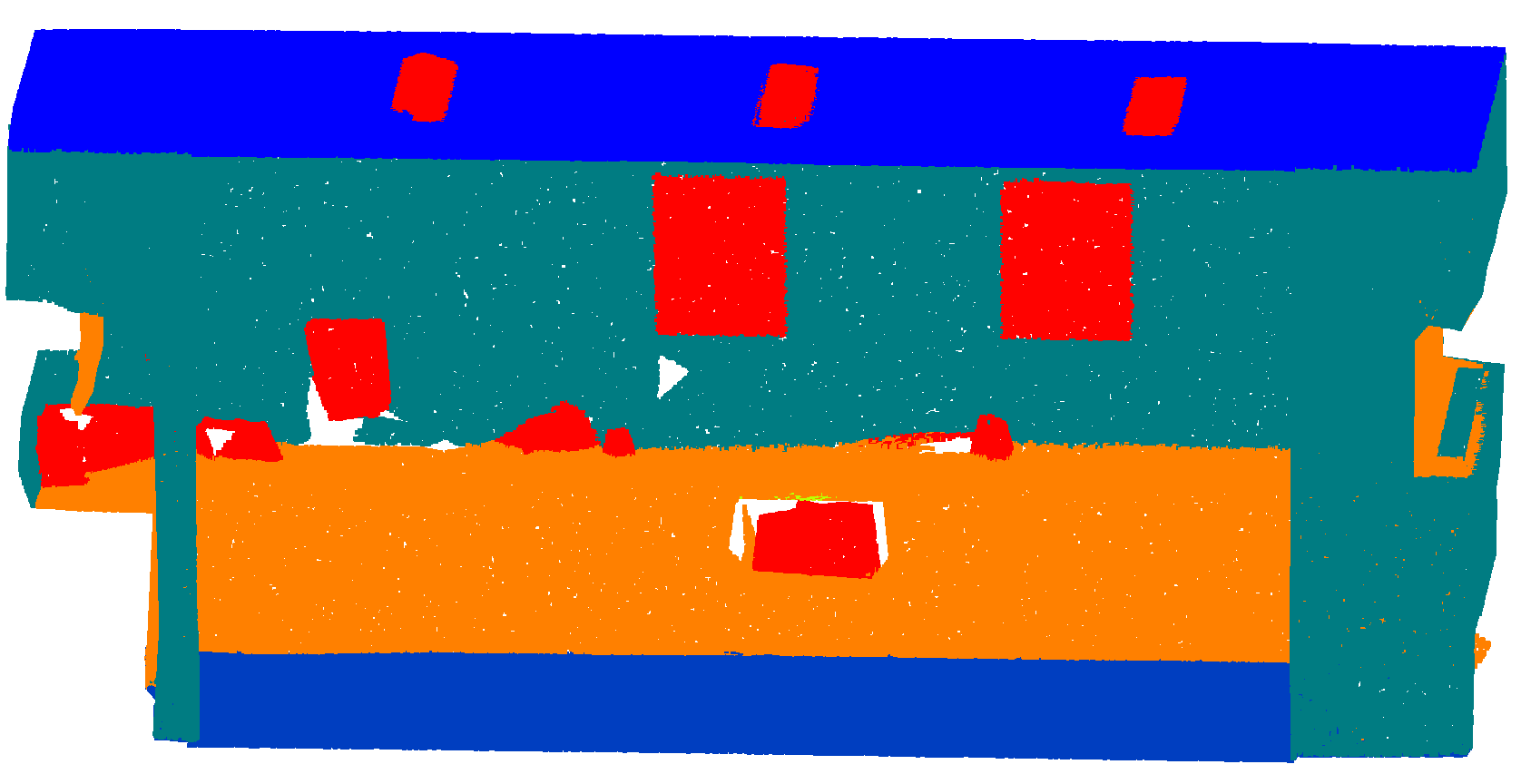}  \\ 
\end{tabular}
\end{table*}

\begin{table*}[ht]
\caption{Qualitative Results on DALES for PT-WNO}
\label{tab:qualitative_dales}
    \centering
\begin{tabular}{ccc} 
Ground Truth & \includegraphics[width=0.45\linewidth]{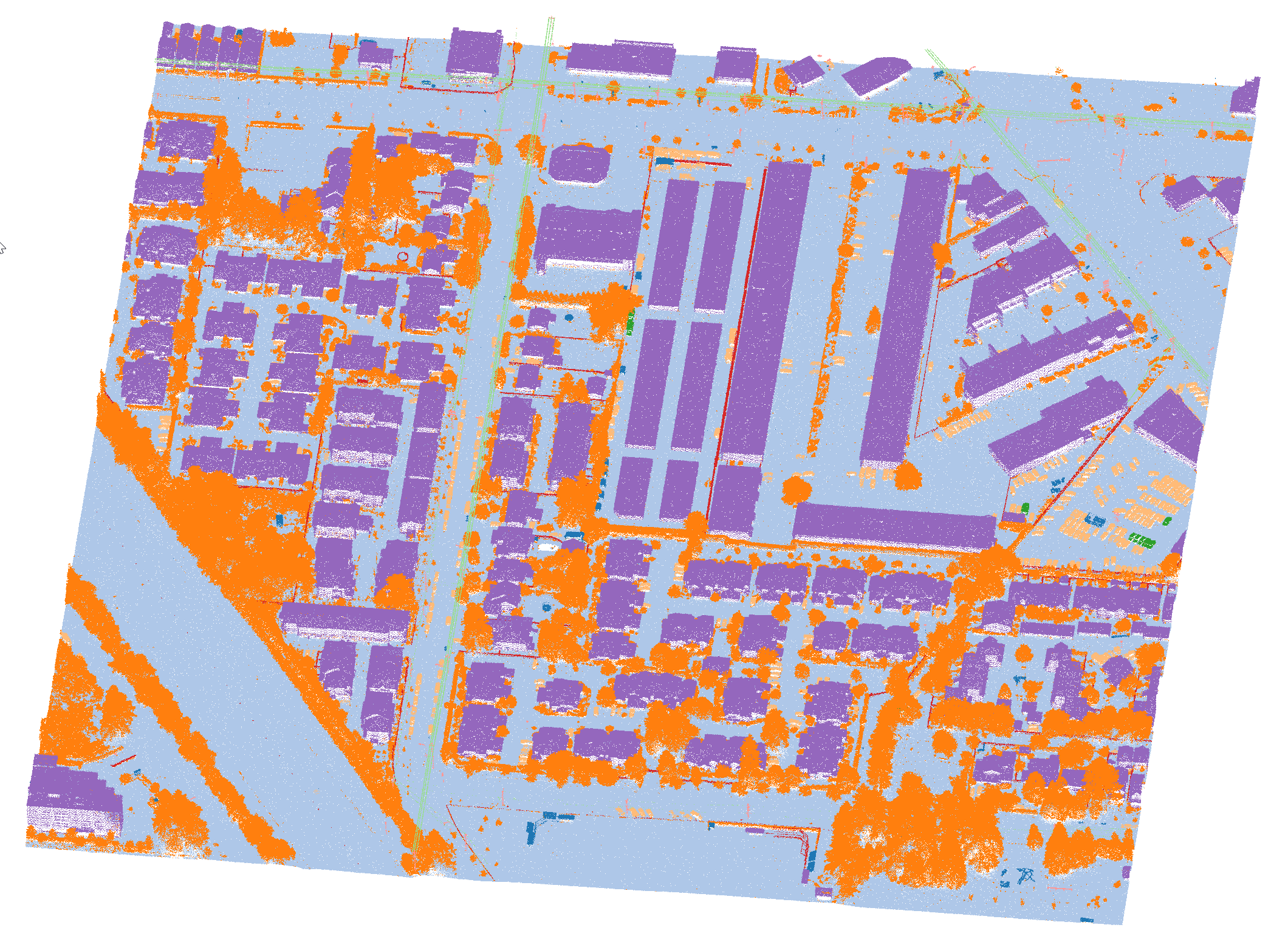}  & \includegraphics[width=0.45\linewidth]{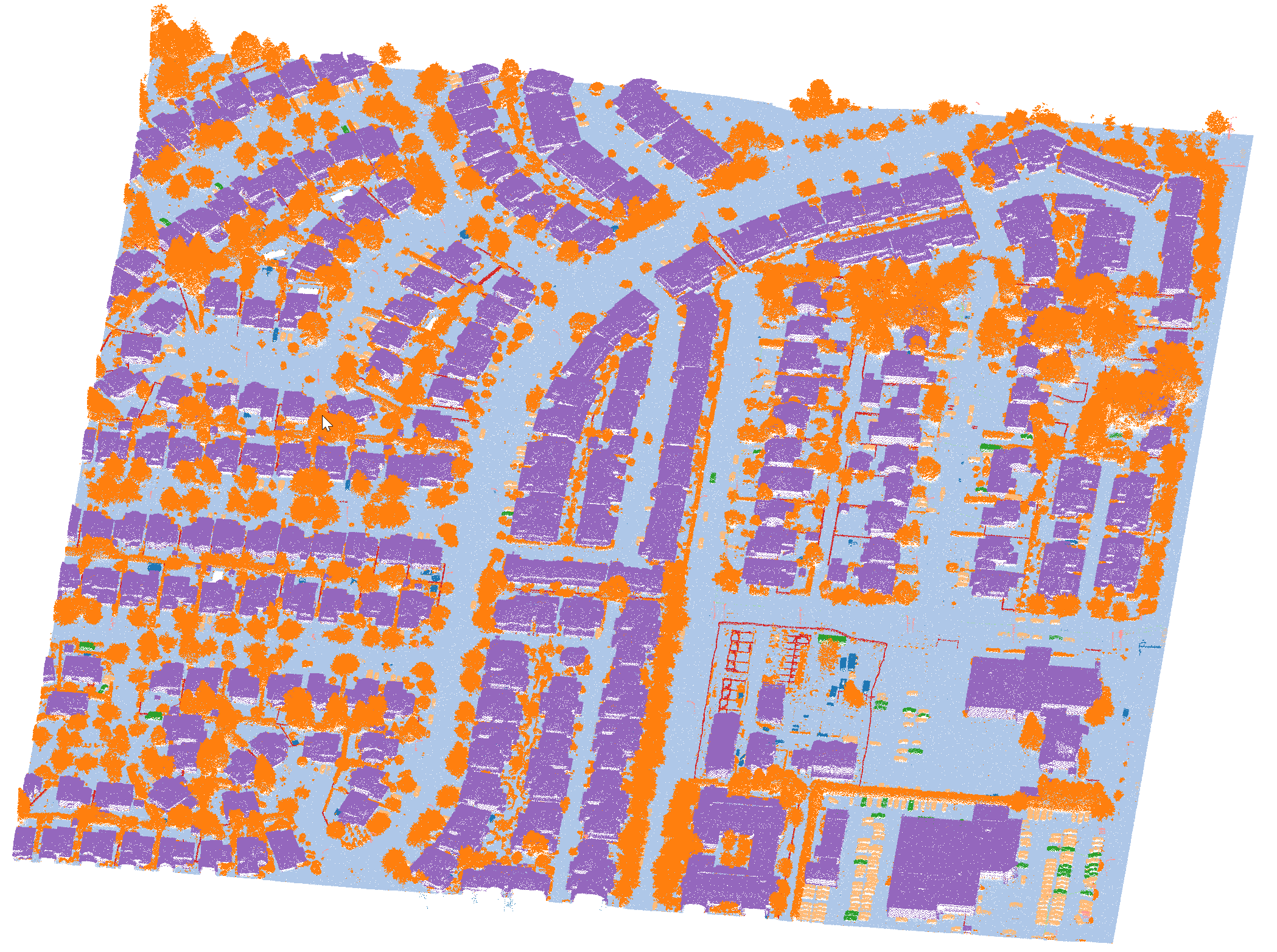} \\
Prediction & \includegraphics[width=0.45\linewidth]{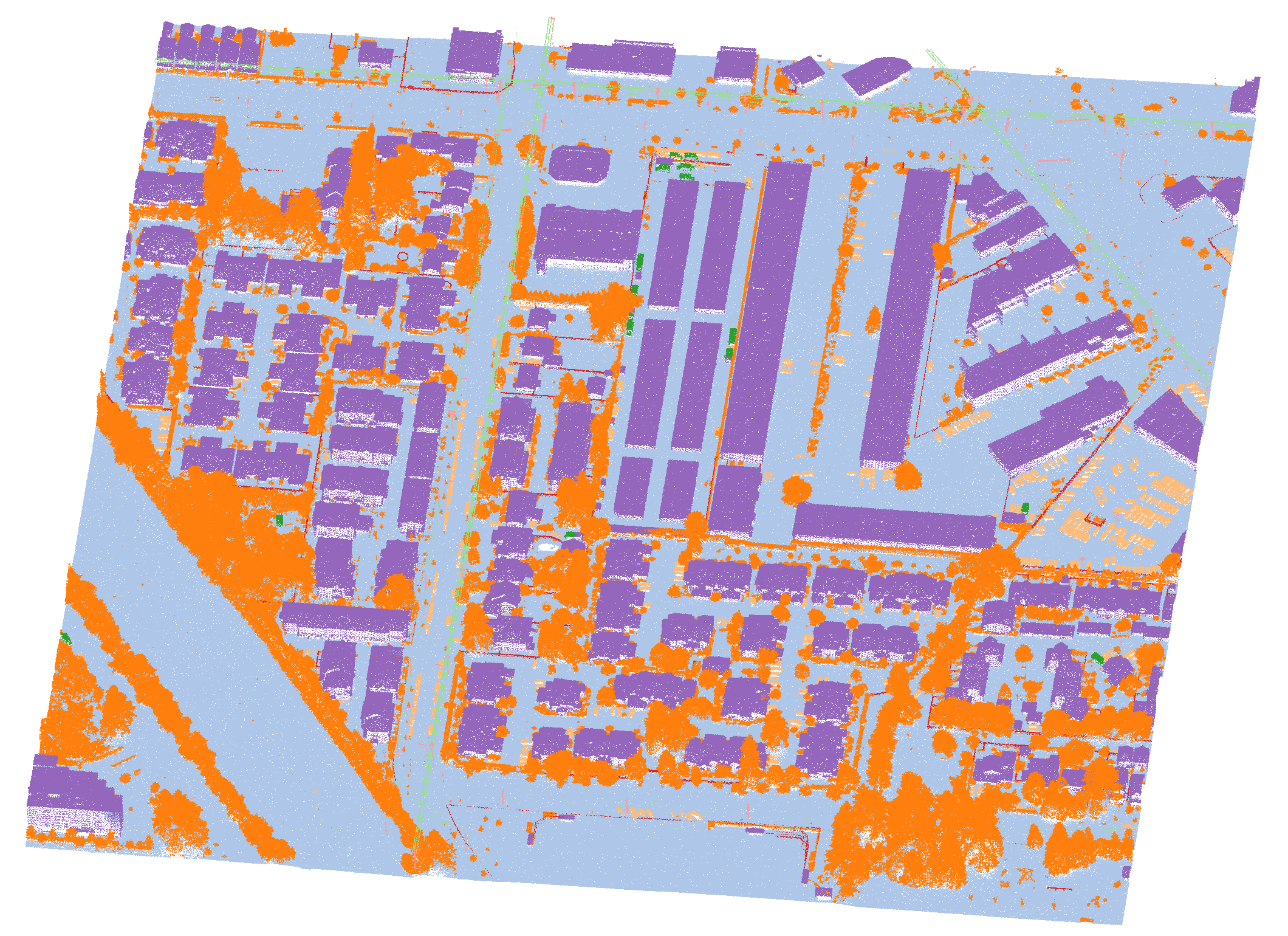}  & \includegraphics[width=0.45\linewidth]{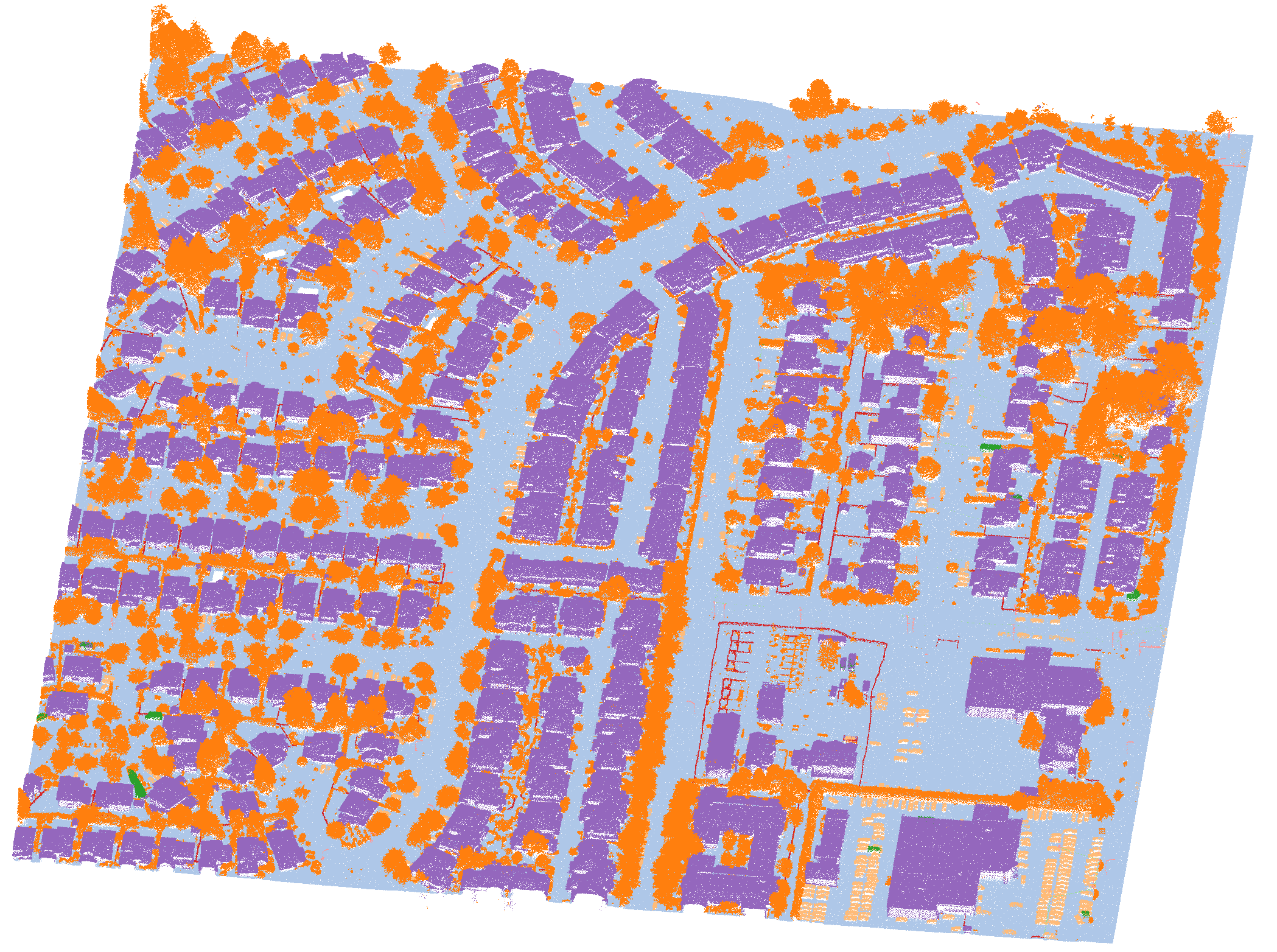} \\ 
\end{tabular}
\end{table*}

\subsection{Experimental Setup}

\subsubsection{Datasets and Metrics.}
For indoor evaluation, we adopt S3DIS~\cite{armeni2016s3dis} (13 classes),
voxelized at $0.02\,\text{m}$ and evaluated on Area 5. For outdoor aerial
evaluation, we use DALES~\cite{varney2020dales} (8 classes), a large-scale aerial
LiDAR dataset spanning $10\,\text{km}^2$ of urban and suburban terrain, voxelized
at $0.05\,\text{m}$ with color augmentations removed to test raw geometric
reasoning. We additionally report ScanNet v2~\cite{dai2017scannet} validation
results (20 classes) to assess transferability to dense mesh-reconstructed scenes.
We report mIoU, mAcc, and OA as evaluation metrics.

\subsubsection{Implementation Details.}
All models are optimized with AdamW (learning rate $6\times10^{-3}$, weight decay
$0.05$), with a reduced learning rate of $6\times10^{-4}$ applied to transformer
attention block parameters. A one-cycle cosine learning rate schedule is used over
3000 training epochs. The WNO branch operates on a spatial grid of resolution
$G_x \times G_y \times G_z = 64^3$ with a universal channel dimension of $D{=}64$.

\subsubsection{Baseline Reproducibility Protocol.}
A key challenge in point cloud benchmarking is that absolute metrics are sensitive
to hardware scale and environment. To ensure a fair comparison, all prior methods
are cited directly from their respective publications. Our reproduced
PTv3~\cite{wu2024ptv3} baseline and all PT-WNO variants are trained entirely from
scratch under identical hardware, hyperparameters, and random seeds, so any
observed performance delta is solely attributable to the wavelet operator.

\subsection{Indoor Scene Segmentation}

Table~\ref{tab:performance} reports overall mIoU across all benchmarks.
On S3DIS Area 5, PT-WNO achieves $91.28\%$ OA, $76.90\%$ mAcc, and $71.59\%$ mIoU, outperforming the controlled PTv3 baseline by $+1.03$ mIoU points with a minimal parameter overhead of $1.6$M ($46.2$M $\rightarrow$ $47.8$M, $+3.5\%$). Table~\ref{tab:s3dis_classwise} details the class-wise breakdown: PT-WNO demonstrates pronounced gains on geometrically distinct classes such as columns ($+4.77$), doors ($+2.40$), and sofas ($+7.21$), where
spatial-frequency localization of WNO captures sharp boundary transitions more effectively than strictly local attention.
\\
On ScanNet v2~\cite{dai2017scannet}, PT-WNO achieves $76.19\%$ mIoU compared to $76.36\%$ for the reproduced PTv3~\cite{wu2024ptv3} baseline, a difference of $-0.17$ points within run-to-run
variance (see Table~\ref{tab:scannet_classwise}). On dense, mesh-reconstructed scans where local attention already has sufficient surface context, the wavelet branch introduces no regression while adding no meaningful gain - consistent with its design as a geometric complement rather than a replacement for local reasoning.

\subsection{Large-Scale Outdoor Segmentation: DALES}

PT-WNO raises mIoU from $79.58\%$ to $81.05\%$ ($+1.47$ points) over the
reproduced PTv3 baseline, approaching the previously strongest published result of KPConv~\cite{Thomas_2019_ICCV} ($81.1\%$) without any hand-crafted convolution kernels. Gains are consistent across geometrically sparse categories: vegetation ($+4.30$), cars ($+2.37$), fences ($+3.12$), and poles ($+3.56$) (see Table~\ref{tab:dales_classwise}). The contrast between DALES~\cite{varney2020dales} ($+1.47$) and ScanNet v2~\cite{dai2017scannet} ($-0.17$) confirms that the wavelet operator's benefit scales with geometric sparsity and the absence of dense color cues.

\subsection{Qualitative Results}
To visually demonstrate the effectiveness of the proposed architecture, we present qualitative segmentation results of PT-WNO on representative scenes from the S3DIS and DALES datasets. These visualizations illustrate the model's capability to generate spatially coherent and structurally accurate predictions across both dense indoor environments and large-scale aerial landscapes. By explicitly aggregating global and multiscale context through the wavelet operator branch, PT-WNO effectively parses complex 3D geometries and maintains semantic continuity over expansive regions.

In the indoor scenes from S3DIS, PT-WNO successfully captures the intricate layouts of large rooms, offices, and extended corridors. The model produces clean, continuous boundaries for major structural elements such as walls, floors, and ceilings, while accurately identifying fine-grained objects like furniture and architectural details. The global context provided by the operator branch ensures that identical semantic categories distributed across different parts of a cluttered room are recognized consistently, preserving the overall structural integrity of the indoor space (see Table~\ref{tab:qualitative_s3dis}).

For the large-scale aerial point clouds in the DALES dataset, the visualizations highlight PT-WNO's robustness in handling outdoor topologies and massive spatial scale variations. The model seamlessly segments extended geographic features, distinguishing clearly between bare earth, vegetation, and man-made structures over broad spatial extents. The joint spatial-frequency localization of the wavelet operator is particularly advantageous in this domain, allowing the network to recognize macro-scale geographic patterns while retaining the localized geometry of smaller structures (see Table~\ref{tab:qualitative_dales}).

Overall, these qualitative examples confirm that the integration of a wavelet neural operator enables PT-WNO to achieve comprehensive scene understanding. The model successfully adapts to the distinct geometric and contextual challenges of both indoor building layouts and outdoor aerial topographies.

\subsection{Ablation Studies}
\label{sec:ablations}

\subsubsection{Input Modality: Color Removal.}
Removing RGB color from S3DIS yields $90.87\%$ OA, $77.21\%$ mAcc, and $70.04\%$ mIoU (see Table~\ref{tab:s3dis_classwise}) - a negligible drop of $0.52$ mIoU versus the full-color PTv3~\cite{wu2024ptv3} baseline ($70.56\%$). This confirms that the WNO branch effectively encodes purely geometric cues, making PT-WNO robust to appearance-poor LiDAR data without architectural modification.

\section{Conclusion}
\label{sec:conclusion}

In this paper, we introduced the Point Transformer with Wavelet Neural Operator (PT-WNO), a novel hybrid architecture that addresses the limitation of local-only context aggregation in 3D semantic segmentation. By integrating a shared wavelet operator branch into a serialization-based transformer backbone, PT-WNO explicitly models scene-level structure and captures long-range multiscale relationships without sacrificing the efficiency of point processing. 

Our extensive experimental evaluation demonstrates that treating large-scale 3D context as a learned global operator yields consistent improvements across diverse environments. Under strictly controlled settings, PT-WNO achieves significant gains on S3DIS Area 5 ($+1.03$ mIoU) and the large-scale aerial DALES benchmark ($+1.47$ mIoU) over the state-of-the-art PTv3 baseline, while maintaining comparable performance on dense indoor mesh scans like ScanNet v2. The benefits of our approach are particularly pronounced on geometrically distinct categories and sparse, appearance-poor LiDAR data, as confirmed by our ablation on color removal. Furthermore, qualitative results illustrate that the wavelet operator branch successfully produces spatially coherent predictions, smoothing out boundaries and reducing isolated misclassifications in both indoor rooms and expansive outdoor topologies.

Looking forward, the success of a geometry-aware universal operator branch opens several promising directions for spatial computing. Future work will explore adapting the wavelet neural operator for dynamic spatio-temporal point clouds (e.g., 4D LiDAR sequences) and investigating hardware-aware pruning of the operator branch to further deploy PT-WNO in real-time edge devices for autonomous navigation and city-scale digital twin generation.

\bibliographystyle{ACM-Reference-Format}
\bibliography{refs}







\end{document}